\pdfoutput=1
\documentclass{article}
\usepackage{ijcai16}

\usepackage{times}
\usepackage{graphicx}
\usepackage{fixltx2e}
\usepackage{float}
\usepackage{amsmath}
\usepackage{booktabs}
\usepackage{multirow}

\DeclareMathOperator*{\argmax}{argmax}

\title{Neural Network Translation Models for Grammatical Error Correction}
\author{
Shamil Chollampatt$^{1}$ \and Kaveh Taghipour$^{2}$ \and Hwee Tou Ng$^{1,2}$ \\
$^{1}$NUS Graduate School for Integrative Sciences and Engineering \\
$^{2}$Department of Computer Science \\
National University of Singapore\\
shamil@u.nus.edu, \{kaveh, nght\}@comp.nus.edu.sg 
}

\begin{document}

\maketitle

\begin{abstract}
Phrase-based statistical machine translation (SMT) systems have previously been used for the task of grammatical error correction (GEC) to achieve state-of-the-art accuracy. The superiority of SMT systems comes from their ability to learn text transformations from erroneous to corrected text, without explicitly modeling error types. However, phrase-based SMT systems suffer from limitations of discrete word representation, linear mapping, and lack of global context. In this paper, we address these limitations by using two different yet complementary neural network models, namely a neural network global lexicon model and a neural network joint model. These neural networks can generalize better by using continuous space representation of words and learn non-linear mappings. Moreover, they can leverage contextual information from the source sentence more effectively. By adding these two components, we achieve statistically significant improvement in accuracy for grammatical error correction over a state-of-the-art GEC system.
\end{abstract}

%%%%%%%%%%%%%%%%%%%%%%%
\section{Introduction}
%%%%%%%%%%%%%%%%%%%%%%

Grammatical error correction (GEC) is a challenging task due to the variability of the type of errors and the syntactic and semantic dependencies of the errors on the surrounding context. Most of the grammatical error correction systems use classification and rule-based approaches for correcting specific error types. However, these systems use several linguistic cues as features. The standard linguistic analysis tools like part-of-speech (POS) taggers and parsers are often trained on well-formed text and perform poorly on ungrammatical text. This introduces further errors and limits the performance of rule-based and classification approaches to GEC. As a consequence,
the phrase-based statistical machine translation (SMT) approach to GEC has gained popularity because of its ability to learn text transformations from erroneous text to correct text from error-corrected parallel corpora without any additional linguistic information. They are also not limited to specific error types. Currently, many state-of-the-art GEC systems are based on SMT or use SMT components for error correction  \cite{susanto2014system,felice-EtAl:2014:W14-17,junczysdowmt-grundkiewicz:2014:W14-17}. In this paper, grammatical error correction includes correcting errors of all types, including word choice errors and collocation errors which constitute a large class of learners' errors. 

We model our GEC system based on the phrase-based SMT approach. However, traditional phrase-based SMT systems treat words and phrases as discrete entities. We take advantage of continuous space representation by adding two neural network components that have been shown to improve SMT systems \cite{ha2015lexical,devlin2014fast}. These neural networks are able to capture non-linear relationships between source and target sentences and can encode contextual information more effectively. Our experiments show that the addition of these two neural networks leads to significant improvements over a strong baseline and outperforms the current state of the art.

%%%%%%%%%%%%%%%%%%%%%%%%%%%%%%%%%%%%
\section{Related Work}
%%%%%%%%%%%%%%%%%%%%%%%%%%%%%%%%%%%%

In the past decade, there has been increasing attention on grammatical error correction in English, mainly due to the growing number of English as Second Language (ESL) learners around the world. The popularity of this problem in natural language processing research grew further through Helping Our Own (HOO) and the CoNLL shared tasks \cite{Dale:2011:HOO:2187681.2187725,dale-anisimoff-narroway:2012:BEA,ng-EtAl:2013:CoNLLST,ng-EtAl:2014:W14-17}.
Most published work in GEC aimed at building specific classifiers for different error types and then use them to build hybrid systems \cite{Dahlmeier:2012:NHS:2390384.2390411,rozovskaya-EtAl:2014:W14-17}. One of the first approaches of using SMT for GEC focused on correction of countability errors of mass nouns (e.g., \textit{many informations}$\rightarrow$\textit{much information}) \cite{brockett-dolan-gamon:2006:COLACL}. They had to use an artificially constructed parallel corpus for training their SMT system. Later, the availability of large-scale error corrected data \cite{mizumoto-EtAl:2011:IJCNLP-2011} further improved SMT-based GEC systems. 

Recently, continuous space representations of words and phrases have been incorporated into SMT systems via neural networks. Specifically, addition of monolingual neural network language models \cite{Bengio:2003:NPL:944919.944966,vaswani-EtAl:2013:EMNLP}, neural network joint models (NNJM) \cite{devlin2014fast}, and neural network global lexicon models (NNGLM) \cite{ha2015lexical} have been shown to be useful for SMT. Neural networks have been previously used for GEC as a language model feature in the classification approach \cite{wu-EtAl:2014:W14-17} and as a classifier for article error correction \cite{sun2015convolutional}. Recently, a neural machine translation approach has been proposed for GEC \cite{zheng:2016:NMT}. This method uses a recurrent neural network to perform sequence-to-sequence mapping from erroneous to well-formed sentences. Additionally, it relies on a post-processing step based on statistical word-based translation models to replace out-of-vocabulary words. In this paper, we investigate the effectiveness of two neural network models, NNGLM and NNJM, in SMT-based GEC. To the best of our knowledge, there is no prior work that uses these two neural network models for SMT-based GEC.

%%%%%%%%%%%%%%%%%%%%%%%%%%%%%%%%%%%%%%%%%%%%%%%%%%%%%%%%%%%%%%%%%%%%%%%%
\section{A Machine Translation Framework for Grammatical Error Correction}
%%%%%%%%%%%%%%%%%%%%%%%%%%%%%%%%%%%%%%%%%%%%%%%%%%%%%%%%%%%%%%%%%%%%%%%%

In this paper, the task of grammatical error correction is formulated as a translation task from the language of `bad' English to the language of `good' English. That is, the source sentence is written by a second language learner and potentially contains grammatical errors, whereas the target sentence is the corrected fluent sentence. We use a phrase-based machine translation framework  \cite{koehn2003statistical} for translation, which employs a log-linear model to find the best translation $T^*$ given a source sentence $S$. The best translation is selected according to the following equation:
\begin{equation*}
T^* = \argmax_{T} P(T|S)  = \argmax_{T}\sum_{i=1}^{N}\lambda_i h_i(T,S)
\end{equation*}
where $N$ is the number of features, $h_i$ and $\lambda_i$ are the $i$\textsuperscript{th} feature function and feature weight, respectively. We make use of the standard features used in phrase-based translation without any reordering, leading to monotone translations. The features can be broadly categorized as translation model and language model features. The translation model in the phrase-based machine translation framework is trained using parallel data, i.e., sentence-aligned erroneous source text and corrected target text.  The translation model is responsible for finding the best transformation of the source sentence to produce the corrected sentence.
On the other hand, the language model is trained on well-formed English text and this ensures the fluency of the corrected text. To find the optimal feature weights ($\lambda$), we use minimum error rate training (MERT), maximizing the $F_{0.5}$ measure on the development set \cite{junczysdowmt-grundkiewicz:2014:W14-17}. The $F_{0.5}$ measure \cite{dahlmeier2012better}, which weights precision twice as much as recall, is the evaluation metric widely used for GEC and was the official evaluation metric adopted in the CoNLL 2014 shared task \cite{ng-EtAl:2014:W14-17}. 

Additionally, we augment the feature set by adding two neural network translation models, namely a neural network global lexicon model \cite{ha2015lexical} and a neural network joint model \cite{devlin2014fast}. These models are described in detail in Sections \ref{sec:nnglm} and \ref{sec:nnjm}.

%%%%%%%%%%%%%%%%%%%%%%%%%%%%%%%%%%%%%%%%%%%%%%
\section{Neural Network Global Lexicon Model }
%%%%%%%%%%%%%%%%%%%%%%%%%%%%%%%%%%%%%%%%%%%%%%%
\label{sec:nnglm}

A global lexicon model is used to predict the presence of words in the corrected output. The model estimates the overall probability of a target hypothesis (i.e., a candidate corrected sentence) given the source sentence, by making use of the probability computed for each word in the hypothesis.  The individual word probabilities can be computed by training density estimation models such as maximum entropy \cite{mauser-hasan-ney:2009:EMNLP} or probabilistic neural networks \cite{ha2015lexical}. Following \cite{ha2015lexical}, we formulate our global lexicon model using a feed-forward neural network. The model and the training algorithm are described below.

\subsection{Model}

The probability of a target hypothesis is computed using the following equation:
\begin{equation}
P(T|S) \approx \prod_{i=1}^{|T|} P(t_i|S)
\label{eqn:nnglm}
\end{equation}
where $S$ and $T$ are the source sentence and the target hypothesis respectively, and $|T|$ denotes the number of words in the target hypothesis. $P(t_i|S)$ is the probability of the target word $t_i$ given the source sentence $S$. $P(t_i|S)$ is the output of the neural network. The architecture of the neural network is shown in Figure \ref{fig:nnglm}. $P(t_i|S)$ is calculated by:

\begin{figure}[t]
  \centering
  \includegraphics[width=0.6\columnwidth]{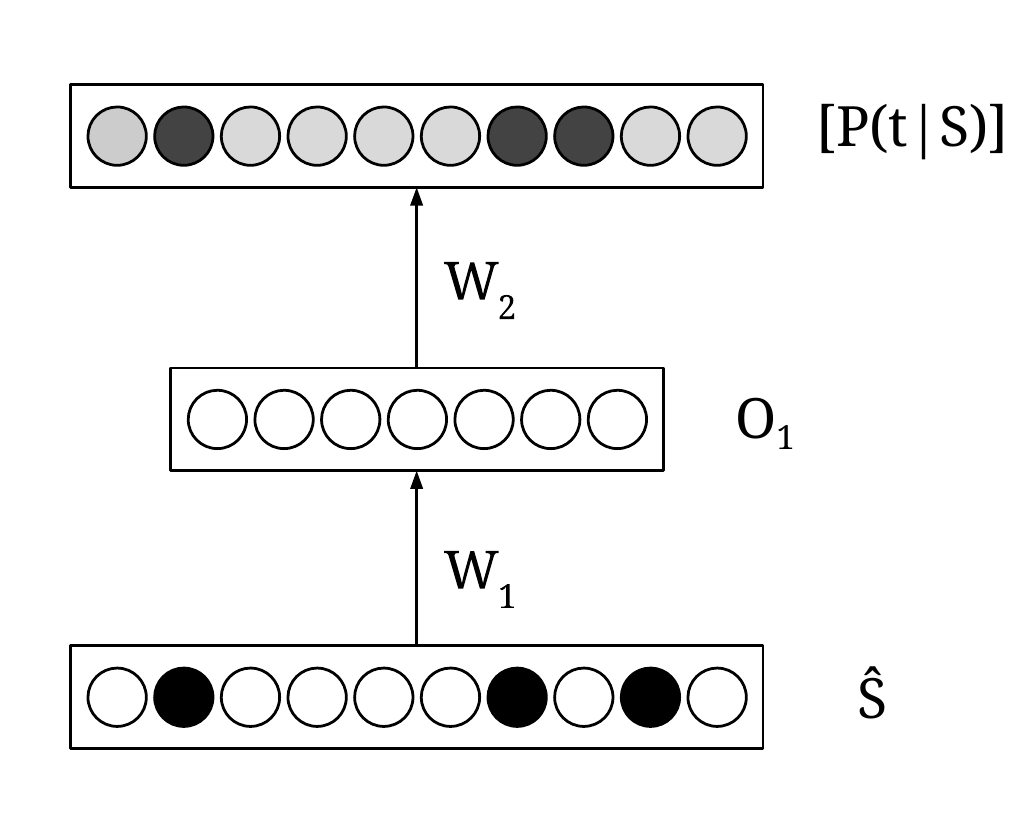}
    \caption{A single hidden layer neural network global lexicon model}
 \label{fig:nnglm}
\end{figure}

\begin{equation*}
P(t_i|S) = \sigma_i(W_2 \cdot O_1 + b_2)
\end{equation*}
 where  $O_1$ is the hidden layer output, and $W_2$ and $b_2$ are the output layer weights and biases respectively. $\sigma_i$ is the element-wise sigmoid function which scales the output to $(0,1)$.
 
 $O_1$ is computed by the following equation:
\begin{equation*}
O_1 = \phi(W_1 \cdot \hat{S} + b_1)
\end{equation*}
where $\phi$ is the activation function, and $W_1$ and $b_1$ are the hidden layer weights and biases applied on a binary bag-of-words representation of the input sentence denoted by $\hat{S}$. The size of $\hat{S}$ is equal to the size of the source vocabulary $|V_s|$ and each element indicates the presence or absence (denoted by 1 or 0 respectively) of a given source word. 

\subsection{Training}

The model is trained using mini-batch gradient descent with back-propagation. We use binary cross entropy (Equation \ref{eq-cross-entropy}) as the cost function:
\begin{align}\label{eq-cross-entropy}
\begin{split}
E = - \dfrac{1}{|V_t|} & \sum_{i=1}^{|V_t|} \left[ \hat{T}_i \log p(t_i|S) \right. \\
	& + \left. (1-\hat{T}_i) \log (1-p(t_i|S)) \right] 
\end{split}
\end{align}
where $\hat{T}$ refers to the binary bag-of-words representation of the reference target sentence, and $V_t$ is the target vocabulary.
Each mini-batch is composed of a fixed number of sentence pairs $(S, T)$. The training algorithm repeatedly minimizes the cost function calculated for a given mini-batch by updating the parameters according to the gradients.

\subsection{Rescaling}
Since the prior probability of observing a particular word in a sentence is usually a small number, the probabilistic output of NNGLM can be biased towards zero. This bias can hurt the performance of our system and therefore, we try to alleviate this problem by rescaling the output \textit{after} training NNGLM. Our solution is to map the output probabilities to a new probability space by fitting a logistic function on the output. Formally, we use Equation \ref{eq-rescaling} as the mapping function:
\begin{equation}
%Q(t_i|S) = 1 / (1 + \exp(-w \cdot P(t_i|S) - b) )
Q(t_i|S) = \dfrac{1}{1 + \exp(-w \cdot P(t_i|S) - b)}
\label{eq-rescaling}
\end{equation}
where $Q(t_i|S)$ is the rescaled probability and $w$ and $b$ are the parameters. For each sentence pair $(S, T)$ in the \textit{development} set, we collect training instances of the form $(x, y)$ for every word $t$ in the target vocabulary, where $x=P(t|S)$ and $y \in \left\{0,1\right\}$. The value of $y$ is set according to the presence ($y=1$) or absence ($y=0$) of the word $t$ in the target sentence $T$. We use \textit{weighted} cross entropy loss function with $L2$-regularization to train $w$ and $b$ on the development set:
\begin{equation*}
E = - \sum_{j=1}^{M} \left[ c_1 y_j \log p(x_j) + c_0 (1-y_j) \log (1-p(x_j)) \right]
\end{equation*}
Here, $M$ is the number of training samples, $p(x)$ is the probability of $x$ computed by $p(x) = 1/(1+\exp(-wx-b))$, and $c_0$ and $c_1$ are the weights assigned to the two classes $y=0$ and $y=1$, respectively. In order to balance the two classes, we weight each class \textit{inversely proportional} to class frequencies in the training data (Equation \ref{eq-class_weights}) to put more weight on the less frequent class:
\begin{equation}
c_0 = M / (2 f_0),\ \ \ \ \ \ c_1 = M / (2 f_1)
\label{eq-class_weights}
\end{equation}
In Equation \ref{eq-class_weights}, $f_0$ and $f_1$ are the number of samples in each class. After training the rescaling model, we use $w$ and $b$ to calculate $Q(t_i|S)$ according to Equation \ref{eq-rescaling}. Finally, we use $Q(t_i|S)$ instead of $P(t_i|S)$ in Equation \ref{eqn:nnglm}.

%%%%%%%%%%%%%%%%%%%%%%%%%%%%%%%%%%%%
\section{Neural Network Joint Model}
%%%%%%%%%%%%%%%%%%%%%%%%%%%%%%%%%%%%
\label{sec:nnjm}

Joint models in translation augment the context information in language models with words from the source sentence. A neural network joint model (NNJM) \cite{devlin2014fast} uses a neural network to model the word probabilities given a context composed of source and target words. NNJM can scale up to large order of n-grams and still perform well because of its ability to capture semantic information through continuous space representations of words and to learn non-linear relationship between source and target words. Unlike the global lexicon model,  NNJM uses a fixed window from the source side and take sequence information of words into consideration in order to estimate the probability of the target word. The model and the training method are described below.

\begin{figure}[t]
  \centering
  \includegraphics[width=\columnwidth]{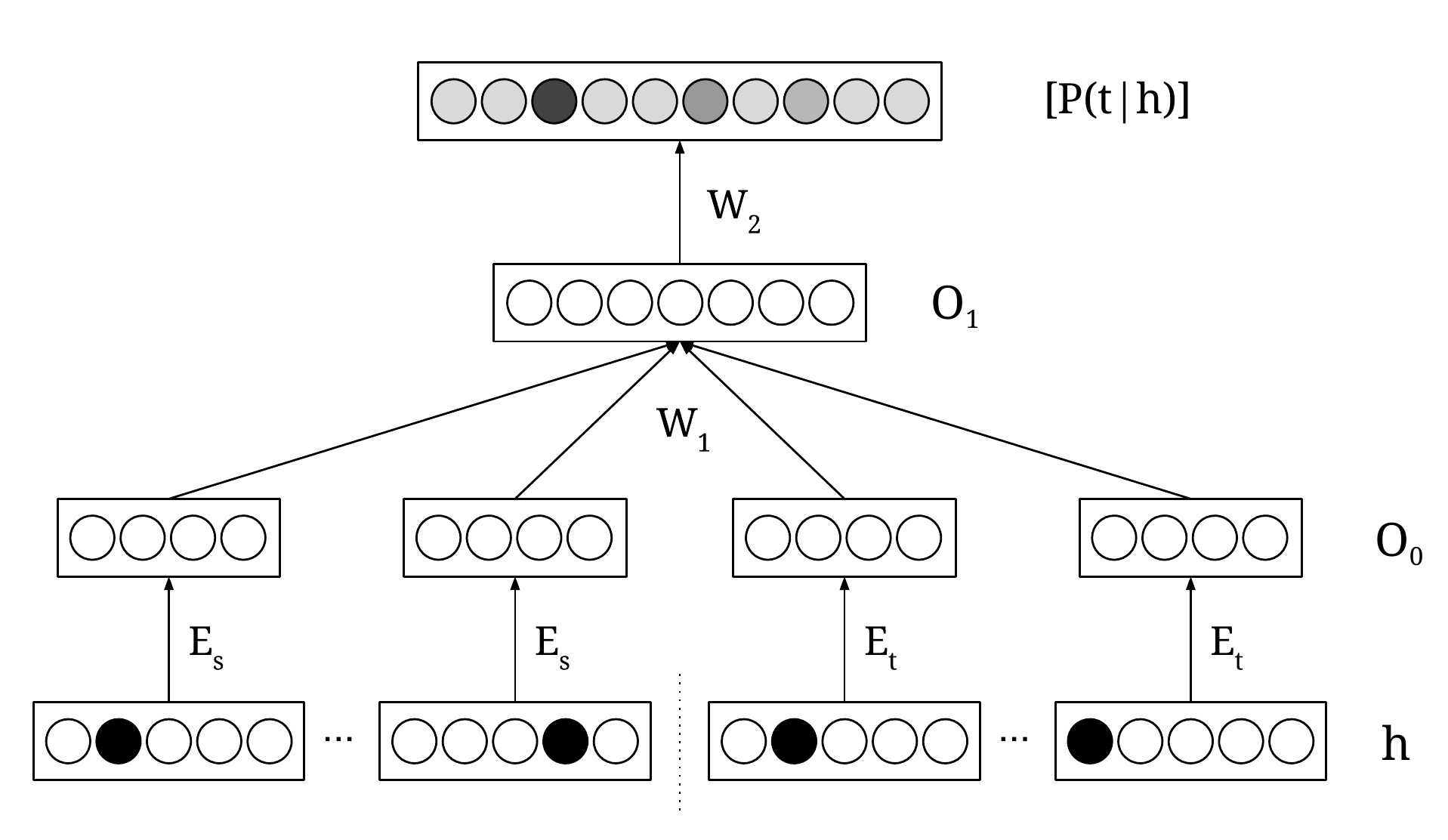}
  \caption{A single hidden layer neural network joint model}
 \label{fig:nnjm}
\end{figure}

\subsection{Model}
The probability of the target hypothesis $T$  given the source sentence $S$ is estimated by the following equation:
\begin{equation}
\label{eqn:nnjm-prob}
P(T|S) \approx \prod_{i=1}^{|T|} P(t_i|h_i)
\end{equation}
where $|T|$ is the number of words in the target sentence, $t_i$ is the $i$\textsuperscript{th} target word, and $h_i$ is the context (history) for the target word $t_i$. The context $h_i$ consists of a set of $m$ source words represented by  $(s_{a_i-\frac{m-1}{2}}, \cdots, s_{a_i+\frac{m-1}{2}})$ and $n-1$ words preceding $t_i$ from the target sentence represented by $(t_{i-n+1}, \cdots, t_{i-1})$. The context words from the source side are the words in the window of size $m$ surrounding the source word $s_{a_i}$ that is aligned to the target word $t_i$. The output of the neural network $P(t_i|h_i)$ is the output of the final softmax layer which is given by the following equation:
\begin{equation}
\label{eqn:nnjm-out}
P(t_i|h_i) = \dfrac{1}{Z(h_i)} \exp U_i(h_i)
\end{equation}
where $U_i(h_i)$ is the output of the neural network before applying softmax and $Z(h_i)$ is given by  following equation:
\begin{equation*}
Z(h_i) = \sum_{i^{'} = 1}^{|V_o|} \exp U_{i^{'}}(h_i)
\end{equation*}
The output of the neural network before softmax is computed by applying output layer weights $W_2$ and biases $b_2$ to the hidden layer output $O_1$. 

\begin{equation*}
U(h_i) = W_2 \cdot O_1 + b_2
\end{equation*}
$O_1$ is computed by applying weights $W_1$ and biases $b_1$ on the hidden layer input $O_0$ and using a non-linear activation function $\phi$:
\begin{equation*}
O_1 = \phi(W_1 \cdot O_0 + b_1)
\end{equation*}
The input to the hidden layer ($O_0$) is a concatenated vector of context word embeddings:
\begin{align*}
O_0 = ( & E_s \cdot \hat{s}_{a_i-\frac{m-1}{2}}, \cdots, E_s \cdot \hat{s}_{a_i+\frac{m-1}{2}}, \\
        & E_t \cdot \hat{t}_{i-n+1}, \cdots, E_t \cdot \hat{t}_{i-1} )
\end{align*}
where $\hat{s}$ and $\hat{t}$ are the \textit{one-hot} representations of the source word $s$ and the target word $t$,
respectively. Similarly, $E_s$ and $E_t$ are the word embeddings matrices for the source words and the target words.

As we use log probabilities instead of raw probabilities in our GEC system, Equation \ref{eqn:nnjm-prob} can be rewritten as the following:
\begin{equation}
\label{eqn:nnjm-logout}
\log P(t_i|h_i) = U_i(h_i) - \log Z(h_i)
\end{equation}
Finally, since the network is trained by Noise Contrastive Estimation (NCE) (described in Section \ref{sec:nnjmtraining}), it becomes self-normalized. This means that $Z(h_i)$ will be approximately 1 and hence the raw output of the neural network $U_i$ can be directly used as the log probabilities during decoding.

\subsection{Training}
\label{sec:nnjmtraining}
To avoid the costly softmax layer and thereby speed up both training and decoding, we use Noise Contrastive Estimation (NCE) following  \cite{vaswani-EtAl:2013:EMNLP}. During training, the negative log likelihood cost function is modified to a probabilistic binary classifier, which learns to discriminate between the actual target word and $k$ random words (noisy samples) per training instance selected from a noise distribution $q$. The two classes are $C = 1$ indicating that the word is the target word and $C=0$ indicating that the word is a noisy sample. The conditional probabilities for $C = 0$ and $C = 1$ given a target word and context is given by:

\begin{equation*}
P(C = 1| t_i, h_i) = \dfrac{\tfrac{1}{k+1} P(t_i|h_i) } {\tfrac{1}{k+1}  P(t_i|h_i) + \tfrac{k}{k+1}  q(t_i) }
\end{equation*}
\begin{equation*}
P(C = 0|t_i, h_i) = \dfrac{ \tfrac{k}{k+1}  q(t_i) } {\tfrac{1}{k+1}  P(t_i|h_i) + \tfrac{k}{k+1}  q(t_i) }
\end{equation*}
where, $P(t_i|h_i)$ is the model probability given in Equation \ref{eqn:nnjm-out}. The negative log likelihood cost function is replaced by the following function. 
\begin{equation*}
L = - \sum_{i} \left[ \log P(C = 1|t_i, h_i) + \sum_{j=1}^{k} \log P(C = 0|\bar{t}_{ij}, h_i) \right]
\end{equation*}
where $\bar{t}_{ij}$ refers to the $j$\textsuperscript{th} noise sample for the target word $t_i$.
$Z(h_i)$ is required for the computation of the neural network output, $P(t_i|h_i)$. However, setting the term $Z(h_i)$ to 1 during training forces the output of the neural network to be self-normalized. Hence, Equation \ref{eqn:nnjm-logout} reduces to: 
\begin{equation}
\label{eqn:nosoftmax}
\log P(t_i|h_i) \approx U_i(h_i)
\end{equation}

Using Equation \ref{eqn:nosoftmax} avoids the expensive softmax computation in the final layer and consequently speeds up decoding.

%%%%%%%%%%%%%%%%%%%%%%%%%%%%%%%%%%%%
\section{Experiments}
%%%%%%%%%%%%%%%%%%%%%%%%%%%%%%%%%%%%
We describe our experimental setup including the description of the data we used, the configuration of our baseline system and the neural network components, and the evaluation method in Section \ref{sec:setup}, followed by the results and discussion in Section \ref{sec:results}

%%%%%%%%%%%%%%%%%%%
\subsection{Setup}
%%%%%%%%%%%%%%%%%%%
\label{sec:setup}
We use the popular phrase-based machine translation toolkit Moses\footnote{https://github.com/moses-smt/mosesdecoder/tree/fscorer} as our baseline SMT system. NUCLE \cite{dahlmeier2013building}, which is the official training data for the CoNLL 2013 and 2014 shared tasks, is used as the parallel text for training. Additionally, we obtain parallel corpora from  \textit{Lang-8 Corpus of Learner English v1.0} \cite{mizumoto-EtAl:2011:IJCNLP-2011}, which consists of texts written by ESL (English as Second Language) learners on the language learning platform Lang-8\footnote{http://lang-8.com/}. We use the test data for the CoNLL 2013 shared task as our development data. The statistics of the training and development data are given in Table \ref{tab:train-data}. Source side refers to the original text written by the ESL learners and target side refers to the corresponding corrected text hand-corrected by humans. The source side and the target side are sentence-aligned and tokenized.

\begin{table}[H]
\centering
\begin{tabular}{@{}lrrr@{}}
\toprule
\textbf{Dataset} & \textbf{\begin{tabular}[c]{@{}l@{}}No. of \\ sentences\end{tabular}} & \textbf{\begin{tabular}[c]{@{}l@{}}No. of source \\ side tokens\end{tabular}} & \textbf{\begin{tabular}[c]{@{}l@{}}No. of target\\ side tokens\end{tabular}}		 \\ \midrule
NUCLE            & 57,151           & 1,161,567                & 1,155,559         \\
\addlinespace[0.2em]
Lang-8 v1.0      & 1,114,139        & 12,945,666               & 13,232,058        \\
\addlinespace[0.2em]
CoNLL 2013       & 1,381            & 29,207                   & 28,743            \\
\bottomrule
\end{tabular}
\caption{Statistics of training and development data}
\label{tab:train-data}
\end{table}

We train the translation model for our SMT system using a concatenation of NUCLE and Lang-8 v1.0 parallel data. The training data is cleaned up by removing sentence pairs in which either the source or the target sentence is empty, or is too long (greater than 80 tokens), or violate a 9:1 sentence ratio limit. The translation model uses the default features in Moses which include the forward and inverse phrase translation probabilities, forward and inverse lexical weights, word penalty, and phrase penalty. We compute the phrase alignments using standard tools in Moses. We use two language model features: a 5-gram language model trained using the target side of NUCLE used for training the translation model and a 5-gram language model trained using English Wikipedia ($\sim$1.78 billion tokens). Both language models are estimated with KenLM\footnote{https://kheafield.com/code/kenlm/} using modified Kneser-Ney smoothing. We use MERT for tuning the feature weights by optimizing the $F_{0.5}$ measure (which weights precision twice as much as recall). This system constitutes our baseline system in Table \ref{tab:results}. Our baseline system uses exactly the same training data as \cite{susanto2014system} for training the translation model and the language model. The difference between our baseline system and the SMT components of \cite{susanto2014system} is that we tune with $F_{0.5}$ instead of BLEU and we use the standard Moses configuration without the Levenshtein distance feature.

On top of our baseline system described above, we incorporate the two neural network components, neural network global lexicon model (NNGLM) and neural network joint model (NNJM) as features. Both NNGLM and NNJM are trained using the parallel data used to train the translation model of our baseline system. 

We implement NNGLM using the Theano library\footnote{http://deeplearning.net/software/theano} in Python in order to make use of parallelization with GPUs, thus speeding up training significantly. We use a source and target vocabulary of 10,000 most frequent words on both sides. We use a single hidden layer neural network with 2,000 hidden nodes. We use $\tanh$ as the activation function for the hidden layer. We optimize the model weights by stochastic gradient descent using a mini-batch size of 100 and a learning rate\footnote{We divide the gradient by the mini-batch size.} of 10. We train the model for 45 epochs. The logistic regression function for rescaling is trained using the probabilities obtained from this model on the development set. To speed up tuning and decoding, we pre-compute the probabilities of target words using the source side sentences of the development and the test sets, respectively. We implement a feature function in Moses to compute the probability of a target hypothesis given the source sentence using the pre-computed probabilities.

To train NNJM, we use the publicly available implementation, Neural Probabilistic Language Model (NPLM) \cite{vaswani-EtAl:2013:EMNLP}. The latest version of Moses can incorporate NNJM trained using NPLM as a feature while decoding.  Similar to NNGLM, we use the parallel text used for training the translation model in order to train NNJM. We use a source context window size of 5 and a target context window size of 4. We select a source context vocabulary of 16,000 most frequent words from the source side. The target context vocabulary and output vocabulary is set to the 32,000 most frequent words. We use a single hidden layer to speed up training and decoding with an input embedding dimension of 192 and 512 hidden layer nodes. We use rectified linear units (ReLU) as the activation function. We train NNJM with noise contrastive estimation with 100 noise samples per training instance, which are obtained from a unigram distribution. The neural network is trained for 30 epochs using stochastic gradient descent optimization with a mini-batch size of 128 and learning rate of 0.1.

We conduct experiments by incorporating NNGLM and NNJM both independently and jointly into our baseline system. The results of our experiments are described in Section \ref{sec:results}. The evaluation is performed similar to the CoNLL 2014 shared task setting using the the official test data of the CoNLL 2014 shared task with annotations from two annotators (without considering alternative annotations suggested by the participating teams). The test dataset consists of 1,312 error-annotated sentences with 30,144 tokens on the source side. We make use of the official scorer for the shared task, M$^2$Scorer v3.2 \cite{dahlmeier2012better}, for evaluation. We perform statistical significance test using one-tailed sign test with bootstrap resampling on 100 samples.

%%%%%%%%%%%%%%%%%%%%%%%%%%%%%%%%%%%%%
\subsection{Results and Discussion}
%%%%%%%%%%%%%%%%%%%%%%%%%%%%%%%%%%%%%
\label{sec:results}

Table \ref{tab:results} presents the results of our experiments with neural network global lexicon model (NNGLM) and neural network joint model (NNJM).

\begin{table}[ht!]
\centering
\begin{tabular}{@{}llll@{}}
\toprule
\textbf{System}                   & \textbf{P}     & \textbf{R}     & \textbf{F$_{0.5}$}   \\ \midrule
Baseline                        & 50.56 & 22.68 & 40.58  \\
Baseline + NNGLM                & 50.73 & 23.21 & \textbf{41.01}\textsuperscript{*} \\
Baseline + NNJM                 & 51.39 & 23.26 & \textbf{41.38}\textsuperscript{*} \\
Baseline + NNGLM + NNJM         & 52.34 & 23.07 & \textbf{41.75}\textsuperscript{*} \\ \bottomrule
\end{tabular}
\caption{Results of our experiments with NNGLM and NNJM on the CoNLL 2014 test set (\textsuperscript{*} indicates statistical significance with $p<0.01$)}
\label{tab:results}
\end{table}

We see that the addition of both NNGLM and NNJM to our baseline individually improves $F_{0.5}$ measure on the CoNLL 2014 test set by 0.43 and 0.80, respectively. Although both improvements over the baseline are statistically significant (with $p<0.01$), we observe that the improvement of NNGLM is slightly lower than that of NNJM. NNGLM encodes the entire lexical information from the source sentence without word ordering information. Hence, it focuses mostly on the choice of words appearing in the output. Many of the words in the source context may not be necessary for ensuring the quality of corrected output. On the other hand, NNJM looks at a smaller window of words in the source side. NNJM can act as a language model and can ensure a fluent translation output compared to NNGLM. 

We also found rescaling to be important for NNGLM because of imbalanced training data. While the most frequent words in the data, `\textit{I}' and \textit{to}', appear in 43\% and 27\% of the training sentences, respectively, most words occur in very few sentences only. For example, the word `\textit{set}' appears in 0.15\% of the sentences and the word `\textit{enterprise}' appears in 0.003\% of the sentences.

%%%%%%%%%%%%EXAMPLES%%%%%%%%%%%%%%%%%%%%%%%%%%%%%
\begin{table*}[ht]
\small
\centering
\begin{tabular}{@{}ll@{}}
\toprule
\multicolumn{2}{c}{\textbf{System: Baseline + NNGLM }}            \\ 
\toprule
\textbf{\textit{Source}}    & However , there are also a great \textbf{amount} of people who against this technology . \\
\textbf{\textit{Baseline}}  & However , there are also a great \textbf{amount} of people who are against this technology . \\
\textbf{\textit{System}}    & However , there are also a great \textbf{number} of people who are against this technology . \\
\textbf{\textit{Reference}} & However , there are also a great \textbf{number} of people who are against this technology . \\
\toprule

\multicolumn{2}{c}{\textbf{System: Baseline + NNJM }}         \\ 
\toprule

\textbf{\textit{Source}}    & The parents give knowledge and love to the children , meanwhile they feel happy with the \textbf{accompany} of the children . \\
\textbf{\textit{Baseline}}  & The parents give knowledge and love to the children , while they feel happy with the \textbf{accompaniment} of the children . \\
\textbf{\textit{System}}    & The parents give knowledge and love to the children , meanwhile they feel happy with the \textbf{company} of the children . \\
\textbf{\textit{Reference}} & The parents give knowledge and love to the children , meanwhile they feel happy with the \textbf{company} of the children . \\
\toprule

\multicolumn{2}{c}{\textbf{System: Baseline + NNGLM + NNJM }}            \\
\toprule

\textbf{\textit{Source}}    & ... by equipping them with a powerful tool to disseminate information almost \textbf{immediate} to the people around them . \\
\textbf{\textit{Baseline}}  & ... by equipping them with a powerful tool to disseminate information almost \textbf{immediate} to the people around them . \\
\textbf{\textit{System}}    & ... by equipping them with a powerful tool to disseminate information almost \textbf{immediately} to the people around them . \\
\textbf{\textit{Reference}} & ... by equipping them with a powerful tool to disseminate information almost \textbf{immediately} to the people around them . \\
\bottomrule
\end{tabular}
\caption{Examples from the outputs of the systems compared against our baseline system. `Source' is the erroneous input sentence, `Baseline' and `System' are the outputs of our baseline and our neural networks-enhanced system, respectively. `Reference' is the corrected sentence in which the corrections are made by a human annotator.}
\label{tab:examples}
\end{table*}
%%%%%%%%%%%%%%%%%%%%%%%%%%%%%%%%%%%%%%%%%%%%%%%%%%%%%%%%%%%
By incorporating both components together, we obtain an improvement of 1.17 in terms of $F_{0.5}$ measure. This indicates that both components are beneficial and complement each other to improve the performance of the baseline system. While NNGLM looks at the entire source sentence and ensures the appropriate choice of words to appear in the output sentence, NNJM encourages the system to choose appropriate corrections that give a fluent output. 

We compare our system to the top 3 systems in the CoNLL 2014 shared task and to the best published results \cite{zheng:2016:NMT,susanto2014system} on the test data of the CoNLL 2014 shared task. The results are summarized in Table \ref{tab:results-conll14}. Our final system including both neural network models outperforms the best system \cite{zheng:2016:NMT} by 1.85 in F$_{0.5}$ measure. It should be noted that this is despite the fact that the system proposed in \cite{zheng:2016:NMT} uses much larger training data than our system.

\begin{table}[ht]
\centering
\begin{tabular}{@{}lccc@{}}
\toprule
\textbf{System}                   & \textbf{P}     & \textbf{R}     & \textbf{F$_{0.5}$}   \\ 
\midrule
Baseline + NNGLM + NNJM         & 52.34 & 23.07 & \textbf{41.75} \\ 
Baseline						& 50.56 & 22.68 & 40.58 		 \\
\midrule
\cite{zheng:2016:NMT}			& - & - & 39.90 \\
\cite{susanto2014system} 		& 53.55	& 19.14	& 	39.39		  \\
\toprule
\multicolumn{4}{l}{\textit{Top 3 systems in the CoNLL 2014 shared task}} \\
\midrule
CAMB	& 	39.71	& 	30.10	& 	37.33 	\\
CUUI	&	41.78	&	24.88	&	36.79	\\
AMU		&	41.62	&	21.40	&	35.01	\\
\bottomrule
\end{tabular}
\caption{Our system compared against competitive grammatical error correction systems. CAMB, CUUI, and AMU are Team IDs in the CoNLL 2014 shared task.}
\label{tab:results-conll14}
\end{table}

We qualitatively analyze the output of our neural network-enhanced systems against the outputs produced by our baseline system. We have included some examples in Table \ref{tab:examples} and the corresponding outputs of the baseline system and the reference sentences. The selected examples show that NNGLM and NNJM choose appropriate words by making use of the surrounding context effectively. 

Note that our neural networks, which rely on fixed source and target vocabulary, map the rare words and misspelled words to the \texttt{UNK} token. Therefore, phrases with the \texttt{UNK} token may get a higher probability than they actually should due to the large number of \texttt{UNK} tokens seen during training. This leads to fewer spelling error corrections compared to the baseline system which does not employ these neural networks. Consider the following example from the test data: \\
\textit{ ... numerous profit-driven companies realize the hugh (\textbf{huge}) human traffic on such social media sites ...}. 
\\
The spelling error \textit{hugh} $\rightarrow$ \textit{huge} is corrected by the baseline system, but not by our final system with the neural networks. This is because the misspelled word \textit{hugh} is not in the neural network vocabulary and so it is mapped to the \texttt{UNK} token. The sentence with the \texttt{UNK} token gets a higher score and hence the system chooses this output over the correct one. 

From our experiments and analysis, we see that NNGLM and NNJM capture contextual information better than regular translation models and language models. This is because they make use of larger source sentence contexts and continuous space representation of words. This enables them to make better predictions compared to traditional translation models and language models. We also observed that our system has an edge over the baseline for correction of word choice and collocation errors.

%%%%%%%%%%%%%%%%%%%%%%%%%%%%%%%%%%%%
\section{Conclusion}
%%%%%%%%%%%%%%%%%%%%%%%%%%%%%%%%%%%%
Our experiments show that using the two neural network translation models improves the performance of a phrase-based SMT approach to GEC. To the best of our knowledge, this is the {\it first} work that uses these two neural network models for SMT-based GEC. The ability of neural networks to model words and phrases in continuous space and capture non-linear relationships enables them to generalize better and make more accurate grammatical error correction. We have achieved state-of-the-art results on the CoNLL 2014 shared task test dataset. This has been done without using any additional training data compared to the best performing systems evaluated on the same dataset.

%%%%%%%%%%%%%%%%%%%%%%%%%%%%%%%%%%%%
\section*{Acknowledgments}
%%%%%%%%%%%%%%%%%%%%%%%%%%%%%%%%%%%%
This research is supported by Singapore Ministry of Education Academic Research Fund Tier 2 grant MOE2013-T2-1-150.

\bibliographystyle{named}
\bibliography{ijcai16}

\end{document}